\documentclass[conference]{IEEEtran}
\IEEEoverridecommandlockouts
\pdfoutput=1
\usepackage{cite}
\usepackage{amsmath,amssymb,amsfonts}
\usepackage{algorithmic}
\usepackage{graphicx}
\usepackage{textcomp}
\usepackage{xcolor}
\usepackage{subcaption}
\usepackage{algorithm}
\usepackage{multirow}
\usepackage{slashbox}
\usepackage[numbers]{natbib}
\usepackage{natbib}

\graphicspath{{Figures/}} 
\usepackage{cancel}
\usepackage{float}
\usepackage{array}
\usepackage[export]{adjustbox}

\def\BibTeX{{\rm B\kern-.05em{\sc i\kern-.025em b}\kern-.08em
    T\kern-.1667em\lower.7ex\hbox{E}\kern-.125emX}}
\begin{document}

\title{Counterfactual Explanation for Auto-Encoder Based Time-Series Anomaly
Detection}

\author{
\IEEEauthorblockN{Abhishek Srinivasan}
\IEEEauthorblockA{\textit{Connected Systems}, 
\textit{Scania CV AB}, S\"odert\"alje, Sweden\\
\textit{KTH Royal Institute of Technology}, Stockholm, Sweden\\
\textit{RISE Research Institutes of Sweden}, Stockholm, Sweden\\
abhishek.srinivasan@scania.com}\\
\and
\IEEEauthorblockN{Varun Singapura Ravi}
\IEEEauthorblockA{\textit{Connected Systems}, 
\textit{Scania CV AB}, S\"odert\"alje, Sweden\\
\textit{Link\"oping University}, Sweden\\
}\\
\and
\IEEEauthorblockN{Juan~Carlos Andresen}
\IEEEauthorblockA{\textit{Connected Systems}, 
\textit{Scania CV AB}, S\"odert\"alje, Sweden\\
}\\
\and
\IEEEauthorblockN{Anders Host}
\IEEEauthorblockA{
\textit{RISE Research Institutes of Sweden}, Stockholm, Sweden\\
\textit{KTH Royal Institute of Technology}, Stockholm, Sweden\\
}\\
}
\maketitle
\begin{abstract}
The complexity of modern electro-mechanical systems require the development of
  sophisticated diagnostic methods like anomaly detection capable of detecting
  deviations. Conventional anomaly detection approaches like signal processing
  and statistical modelling often struggle to effectively handle the intricacies
  of complex systems, particularly when dealing with multi-variate signals. In
  contrast, neural network-based anomaly detection methods, especially
  Auto-Encoders, have emerged as a compelling alternative, demonstrating
  remarkable performance. However, Auto-Encoders exhibit inherent opaqueness in
  their decision-making processes, hindering their practical implementation at
  scale. Addressing this opacity is essential for enhancing the interpretability
  and trustworthiness of anomaly detection models. In this work, we address this
  challenge by employing a feature selector to select features and
  counterfactual explanations to give a context to the model output.  We tested
  this approach on the SKAB benchmark dataset and an industrial time-series
  dataset. The gradient based counterfactual explanation approach was evaluated
  via validity, sparsity and distance measures. Our experimental findings
  illustrate that our proposed counterfactual approach can offer meaningful and
  valuable insights into the model decision-making process, by explaining fewer
  signals compared to conventional approaches.  These insights enhance the
  trustworthiness and interpretability of anomaly detection models.
\end{abstract}

\section{Introduction}
Modern electrical and mechanical systems are increasingly equipped with more
sensors, enabling the development of new  anomaly detection methods to identify
and alert on deviations indicating failures or malfunctioning. Traditionally,
these anomaly detection systems were meticulously designed for specific machines
and specific components. However, this requires deep domain knowledge and
understanding of the systems. 

Recent data-driven approaches offer a compelling alternative. They leverage
generalizable algorithms that can learn from data, eliminating the need for
expert-crafted rules for each specific scenario. This reduces the efforts
required for building an anomaly detector. Neural networks, in particular, have
shown remarkable effectiveness in anomaly detection for various applications
\cite{schmidl2022anomaly}.

Detecting anomalies in a system using sensor data is a task within the field of
multivariate time-series analysis. Current trends of neural-network based
time-series anomaly detection methods fall under two main categories, i.e.,
forecast and reconstruction \cite{schmidl2022anomaly}. The forecasting methods
are state-based models, they learn the inherent mechanism for forecasting the
future states. When the observations and model forecast deviate by a certain
threshold an alarm is raised. On the other hand, the reconstruction-based methods
learn to compress the normal data (fault free) to a lower dimensional latent
space. This lower dimensional latent space is transformed back to the original
space. Any data with the reconstruction error higher than a given threshold is
considered anomalous. 

In real settings just raising anomaly alert is not enough to act upon it.
A context is required, such as to know why the model is flagging an
anomaly and which sensor data is behaving anomalous. Neural networks are
inherently black-box models and neural-network-based anomaly detection does not
naturally provide its internal decision-making process. Significant progress has
been done within the field of explainability in this
direction~\cite{molnar2020interpretable}.  The explainability methods can
provide global or local explanations. The global explanations aim to distill the
model in an easily understandable logic form (i.e., to explain the model mechanism).
Whereas the local explanations aim to explain the prediction of each input
sample, e.g., Saliency map and counterfactuals.

Counterfactual explanation is a promising tool that provides context to the anomalies
found by neural-network-based models. This explanation method is especially
interesting for diagnostic applications, as their explanation focuses on
answering the question: `why is sample A classified as an anomaly and not
normal?'. The usual approach for building counterfactual explanations is to
start from an anomalous sample and optimise it via a cost function, towards a
counterfactual sample which would be classified as normal by the same model that
classified it as anomalous. To our knowledge, there is very limited amount of
work focused on explaining time series anomaly detection
\cite{haldar2021reliable, sulem2022diverse}. From the perspective of
component diagnostics and maintenance, the existing approaches have a crucial
limitation: they often modify all features within a time series to explain the
anomaly. The freedom of adjusting just any signal of the anomalous sample in the
optimisation process to change the classification averages out valuable 
information and spreads it over many signals. This loss of information makes 
it more difficult to interpret the generated counterfactual and makes it less 
useful for root-cause analysis and diagnostics.

For gaining valuable insights into the anomalies, it is crucial to know the specific
features responsible for the anomaly \textit{and} the reason behind the model's
classification. As discussed in the previous paragraph, conventional
counterfactual explanations solely address the reason behind the anomalies. 
In this work, we propose an explanation method that identifies the relevant features \textit{and} simultaneously explains
the reason behind the anomaly detection for time series reconstruction-based models.

Our approach was tested on the SKAB benchmark data~\cite{skab} and on a
real-world industrial time-series data using Auto-Encoder based anomaly
detection. The results show that counterfactual explanations, using the 
proposed approach, provide insightful explanations about the nature of 
the anomalies such as correlation loss and data drift. 

\section{Related Work}

Counterfactual explanation approaches in general have different focuses,
including generating valid, sparse, actionable, and causal explanations
\cite{Verma2020CounterfactualEF}. Few address the problem of explaining
time-series or anomaly detection. The authors of~\cite{haldar2021reliable}
investigate the challenge of generating robust
counterfactuals for anomaly detection. They define robust counterfactuals as
counterfactual samples that don't flip back to the original class in the
vicinity of a certain distance. They solve this by adding a constraint in the
cost function used for counterfactual optimisation.
In~\cite{sulem2022diverse} the authors build upon the
previous work DiCE \cite{mothilal2020explaining} for generating diverse
counterfactual bounds for time-series anomaly detection. They promote diversity
on the generated counterfactual to address the problems of classical
counterfactual explanation methods, i.e., generating only one of many possible
solutions.  Here, their focus was to provide explanation bounds through diverse
explanations. 

Other research \cite{li2023survey,li2023survey,antwarg2021explaining} utilise 
feature importance, a different class
of explanations, for Auto-Encoder based anomaly detection. In contrast to ours,
their studies do not target time-series data. The authors of \cite{antwarg2021explaining}
use a Shapley-values-based approach (feature
importance) for Auto-Encoders to explain the impact of a certain feature on
other features reconstruction. \cite{chakraborttii2020improving} use feature
level thresholds for explanations and use feature selection to raise alarms
individually. However, they do not explain the reason behind the model
prediction.

To our knowledge, previous work has focused on providing either the relevant
features or the reason behind anomaly detection. Whereas our approach provides
both; the relevant features responsible for the anomaly and the reason why the
model classified it as an anomaly. These two factors play a vital role in 
planing a meaningful action for diagnostics,  such as troubleshooting and 
maintenance scheduling.

\section{Preliminaries}
\subsection{Auto-Encoder (AE)}
Auto-Encoders (AE) are unsupervised modeling approaches. An AE model reduces the
input, i.e., high dimensional data $x \in \mathbb{R}^n$ into a low dimensional
latent representation (encoding) $z \in \mathbb{R}^k$, where $k<n$, using an
encoder $E(x, w_e)$. This encoder is followed by a decoder $D(z, w_d)$ which
reconstructs the input (decoding) $\hat{x} \in \mathbb{R}^n$ from the latent
representation. The encoder and decoder are neural networks with parameters
$w_e$ and $w_d$, respectively. The training process optimises the parameters of
the encoding and decoding functions to provide a reconstruction $\hat{x}$ as
close as possible to the input $x$.  Some common loss functions utilised are
mean square error (MSE), mean absolute error (MAE), and Huber loss.

To extend the AE approach to time-series data we use convo-lution-based
architectures for the encoder and the decoder. We pre-process the data into
time-windows. A time-window of length $l$ is represented as $X = \left(x_{t},
..., x_{t+l}\right)\in \mathbb{R}^{n\times l}$, where $x_t\in \mathbb{R}^n$ are
the signal values at time $t$.

\subsection{Gradient based Counterfactual Explainer} 
\label{section:explanation} 
In this section, we outline the fundamental principles of gradient
based counterfactual explanation techniques. Counterfactual explanations are
generated by gradient optimisation on the objective function~\cite{wachter2017counterfactual}. The
objective function $l(x')$ written in general form is given by

\begin{equation} 
l(x') = cost(x', model(x')) + (\lambda * d(x,x')) \;, \label{eq: couter_cost} 
\end{equation} 
where $x$ is the sample, $x'$ is the generated
counterfactual, $\lambda$ is the weighted factor and the function $d(., .)$ is a
distance measure. This objective function contains two parts, the first part
optimises to flip the class (from anomalous to non-anomalous) of the provided
anomalous sample and the second minimizes the change between the explanation and
the provided sample. Other custom parts can be added depending on the use-case. 

In addition to requiring an objective function, this approach also requires the
model to be differentiable to be able to use a gradient-based optimisation for
counterfactual generation. A simple gradient descent optimisation is given by
 
\begin{equation} 
x'_{i} = x'_{i-1} - \eta . \nabla l(x'_{i-1})\,, \label{eq: opt} 
\end{equation} where $i$ is the optimisation iteration number, $\eta$ is
the step length and $x'_{i-1}$ is the sample form the previous iteration. 

\section{Method}
\label{sec: methods}
Our approach has three different modules; illustrated in figure \ref{fig:
methods illustration}: 1) Anomaly detector, 2) Feature selector, and 3)
Counterfactual explainer. The anomaly detector detects the anomalies. If the
provided sample is anomalous, the feature selector provides a list of relevant
features to be explained. The counterfactual explainer builds an explanation on
the relevant signals that the feature selector selects. 

\begin{figure}
\includegraphics[width=0.47\textwidth]{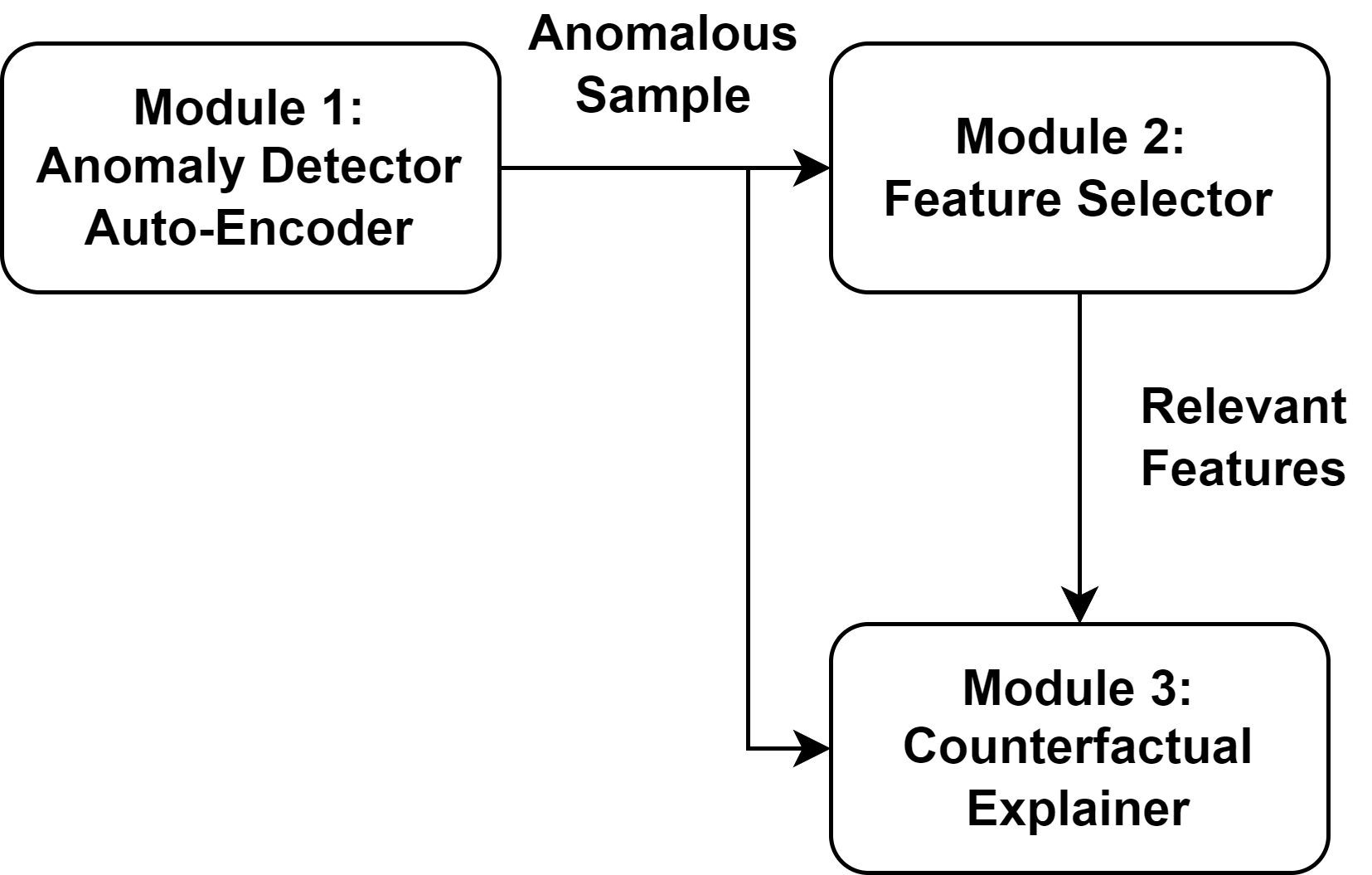}
\caption{
  Our proposed methods has 3 modules, 1) Anomaly detector, 2) Feature
  selector, and 3) Counterfactual explainer. The samples that are classified
  anomalous by the anomaly detector (module 1) are explained though the feature
  selector (module 2) and the counterfactual explainer (module 3). The explainer
  (module 3) uses the selected features from the feature selector and the input
  sample.  }
  \label{fig: methods illustration}

\end{figure}

The anomaly detector (module 1) uses an AE, with an encoder \textit{E} and a
decoder \textit{D}. The encoder \textit{E} consists of 1D-convolution layers
followed by fully connected layers, where-as the decoder \textit{D} uses a
mirrored architecture starting with fully connected layers and then 1-D
transpose convolution layers. The resulting outputs from the decoder have the
same dimension as the inputs. The AE is trained to minimize the reconstruction
loss using Huber loss given by

\begin{equation} 
L(Y) = \frac{1}{M}\sum_{ij} \begin{cases} 0.5 \cdot y_{ij}
/\beta, & \text{for } \sqrt{y_{ij}} < \beta \\ 
\sqrt{y_{ij}} - 0.5 \cdot\beta, & \text{otherwise} \end{cases}\,, 
\label{eq: huber} 
\end{equation}

where $Y=(X-\hat{X})^2_{\circ} \in \mathbb{R}^{n\times l}$, the ${\circ}$
denotes element wise operation, $M$ is the number of elements of the matrix
$Y$, $X$ is the input to AE and $\hat{X}$ is the reconstruction from AE. Once
the AE is trained, the anomaly score (AS) for the validation set is calculated
using 

\begin{equation} 
AS(X, \hat{X}) =  MSE(X,\hat{X}) + MAE(X,\hat{X})\,,\label{eq: ano_scr} 
\end{equation} 

where $\left\{X,\hat{X}\right\}\in \mathbb{R}^{n\times l}$ and  $MSE(\cdot,
\cdot)$ is the mean square error and $MAE(\cdot, \cdot)$ is the mean absolute
error of all elements of the matrices. The mean squared error (MSE) element
emphasizes larger errors (greater than one) more heavily than the mean absolute
error (MAE).  Conversely, MAE penalizes smaller errors (below one) more
severely. This combination of properties contributes to the effectiveness of the
AS. Scores above a threshold are considered anomalous, where the
threshold is defined as $ \theta_{th} = \mu_{ scr} + (k * \sigma_{ scr})$ and
$\mu_{ scr}$ is the mean anomaly score on the validation set, $\sigma_{ scr}$
is the standard deviation of anomaly scores on the validation set and $k$ a parameter.

Explanations are provided by the next two modules only when a given sample is
classified as anomalous, i.e., when the AS is above the defined $\theta_{th}$.
The feature selector (module 2) selects features relevant to the anomaly. It 
processes the anomalous time window and identifies the features as having 
either a high or low impact on the anomaly.
High-impact features are defined as the ones that are
over $m \times percentile(ASW, 90)$ for more than $90\%$ for the window
duration, where we choose $m=0.75$ and $ASW$ is the anomaly score for each
feature and time point in the window and is given by \begin{equation} ASW(X,
\hat{X}) =  (X - \hat{X})^2_{\circ} + |X - \hat{X}|_{\circ} \,, \label{eq:
ano_scr_window} \end{equation} where $\left\{X,\hat{X}\right\}\in
\mathbb{R}^{n\times l}$ and the $\circ$ denotes element wise operation.  The key
difference between equation~(\ref{eq: ano_scr}) and equation~(\ref{eq:
ano_scr_window}) lies in the averaging of the error term. ASW in
Equation~(\ref{eq: ano_scr_window}) does not average the error, retaining the
time and feature dimension assists feature selector to select the right features
where anomalies are observed.
 
The counterfactual explainer (module 3) takes in an anomalous time-window and
the features selected by the feature selector. The counterfactual generator uses
a modified gradient based explanation (see section \ref{section:explanation}).
The difference is that the counterfactual explanation is generated only for the
selected features by module 2. This is done by setting the gradients of
non-selected features to zero and using the same equation~(\ref{eq:  opt}) for
optimisation, where the $cost$ term is given by the AS in 
equation~(\ref{eq: ano_scr}) and the $model$ given by the anomaly detection 
AE model.
  
\subsection{Evaluation Metrics}
\subsubsection{Anomaly Detection Evaluation}
As a sanity check, the developed anomaly detection is evaluated with three 
different metrics; F1-score, False Positive Rate (FPR) and Recall. Equations 
for these evaluation measures are provided by

\begin{equation}
    \label{eq: F1_score}
    \text{F1-score} = \frac{TP}{TP + (0.5*(FP+FN))}\,,
\end{equation}

\begin{equation}
    \label{eq: FPR}
    \text{FPR} = \frac{FP}{FP + TN}\,,
\end{equation}

\begin{equation}
    \label{eq: recall}
    \text{Recall} = \frac{TP}{TP+FN}\,,
\end{equation}
where, TP, FP, TN, and FN refer to true positive, false positive, true 
negative, and false negative, respectively.

\subsubsection{Explainability Evaluation}
\label{sec: explainability eval}
The developed explainability approach is evaluated with measures:
\textit{validity, sparsity, and distance}. \textit{Validity} checks if the
generated counterfactual is valid, i.e., if the produced counterfactual is
classified as normal.  \textit{Sparsity} measures the proportion of features
changed in order to generate the counterfactual. Finally, the \textit{distance}
provides the mean absolute error distance between the sample and counterfactual.

\begin{equation}
	\textit{validity}(x') = \frac{1}{N} \sum_{i=1}^{N} \chi 
    \left(AS(x_{i}', AE(x_{i} ') ) < \theta_{th} \right)\,,
\end{equation}

\begin{equation*}
	ind(x, x') = \chi(\frac{1}{l}(\sum_{j=1}^{l} |x_{ijk} - x_{ijk} '|) > \epsilon) \,,
\end{equation*}

\begin{equation}
	sparsity(x, x') = \frac{1}{N} \sum_{i=1}^{N} \left( \frac{1}{n}  
    \sum_{k=1}^{n} ind(x_{ijk}, x_{ijk}') \right)\,,
\end{equation}

\begin{equation}
	d(x,x') = \frac{1}{N} \sum_{i=1}^{N} \left( \frac{1}{l 
    \cdot n} \sum_{j,k}  |(x_{ijk} - x_{ijk}')| \right)\,,
\end{equation}
where $\left\{x, x'\right\}\in \mathbb{R}^{N \times n\times l}$
\begin{itemize}
	\item $N$: the number of samples,
	\item $l$: the sequence length, i.e., the number of time steps per sequence,
	\item $n$: the number of features,
	\item $x$: sample to be explained, 
	\item $x'$: the counterfactual explanation, 
        \item $\theta_{th}$ the threshold used for anomaly detector,
	\item $\epsilon$: limit defining significant change.
	\item $\chi(c)$: the indicator function, returning 1 when its argument 
        condition $c$ is true, and 0 otherwise.
    \item $AE(c)$: is the Auto-Encoder model.
\end{itemize}
The significant change $\epsilon$ in sparsity allows some wiggle room. Typically, 
this parameter is defined based on the context and the application. In this study 
$\epsilon$ is set to $0.005$, i.e., any change above is counted to be a significant.

\section{Experimental Setting}
\subsection{SKAB dataset}
\cite{skab} designed a benchmark dataset for time-series anomaly detection. This
data is collected from a test-rig consisting of a water tank, valves, and a
pump. In this setup, the pump is specifically crafted to extract water from the
tank and subsequently circulate it back into the same tank. This setup is
equipped with numerous sensors like accelerometer on the pump, pressure sensor
after the pump, thermocouple in water, current, and voltage, in total of 8
signals. The collected data is organised in four parts 'no faults', 'valve 1',
'valve 2', and 'others'. 'No fault' has data from normal operation. Data in
'valve 1' and 'valve 2' has data where the corresponding valves were closed for
partial duration. The 'others' comprises data from multiple anomaly categories
including rotor imbalance, cavitation, and fluid leaks. Each file in 'valve 1',
'valve 2' and 'others' is part normal and part anomalous. It is crucial to note
that no two anomaly types co-occur at the same time.  The data utilization from
different parts of the dataset is summarised in the table \ref{tab: data_usage}.
The files $1-4$ are omitted as the data is marked to be simulated and has
different characteristics than the other files. After pre-processing into
windows, the size of train, validation and test set is  18584, 4658 and 10426
samples. Out of  10426 test samples 3876 are anomalies. 

\begin{table}[h]
    \centering
    \scalebox{0.9}{ 
    \begin{tabular}{c||c|c}
        Dataset & Used as & Files \\
        \hline
        \hline
         Anomaly-free & 80\% Train, 20\% Valid & All\\
         Valve 1 & 80\% Train, 20\% Valid & Only normal behaviour\\
         Valve 2 & 80\% Train, 20\% Valid & Only normal behaviour\\
         Others & Test & 5-14\\
    \end{tabular}
    }
    \caption{Table summarizing utilization of SKAB dataset used in our experiments.}
    \label{tab: data_usage}
\end{table}

\subsection{Real-world industrial Data}
A commercial, real-world industrial data was collected from a field truck. This
data consists of recordings from sensors during normal and anomalous behaviour.
Similar to SKAB data, this industrial data encompasses two anomaly types, with
no instances of simultaneous occurrences. Two different anomalies were
considered: ``correlation loss'' and ``change in relation''. A set of 11
relevant sensor signals were utilised for the experiment. The training and
validation processes were conducted using two separate dataset containing only
normal data (i.e., no-fault data). The test set involved one no-fault scenario
and two anomalous runs, where the anomalies were of a different nature. After
pre-processing into windows the number of samples in train, validation and test
set is 3231, 1074 and 4355 samples. Out of  4355 test samples 1396 were
anomalies.

\subsection{Model and Explainer Setup}
To pre-process the data, we have used min-max normalisation. This involves 
using the minimum and maximum values from the train-set to normalise the 
train, validation, and test sets. The time-series sensor signals were pre-processed 
into smaller chunks using a sliding window technique, with a window length \textit{l} 
of 64 over \textit{n} signals, \textit{n} being 8 and 11 for SKAB and real-world 
data respectively. 

Experiments on the SKAB dataset employed a random seed of 125. The AE model 
consists of: i) Encoder with 2 layers of 1D convolution with 64 and 32 filters, 
kernel size of 5 and stride of 2, followed by a fully connected layer of 8 units; 
ii) Decoder consists of a mirrored architecture to the above, starting with a 
dense layer of size 128 followed by 2 layers of 1D transpose convolution with 
32 and 8 filters, kernel size of 5 and stride of 2. The model was trained for 
150 epochs with a batch size of 64, using Adam optimiser with a learning rate 
of $\lambda = 0.001$, parameters $\beta1 = 0.9$, and $\beta2 = 0.999$, we set
$k=8$ for calculating $\theta_{th}$.

Experiments on the industrial employed uses a random seed of 42. The AE model 
consists of: i) Encoder with 2 layers of 1D convolution with 32 and 64 filters, 
padding 1, kernel size of 5 and stride of 1, followed by 4 fully connected layers 
with  64, 32, 16, and 8 units; ii) Decoder consists of the mirrored architecture, 
starting with 2 dense layers of size 16 and 32, followed by 2 layers of 1D 
transpose convolution with 64 and 32 filters, kernel size of 5 and stride of 1. 
The model was trained for 100 epochs with a batch size of 32, using Adam 
optimiser (AMSGrad variant) with a learning rate of $\lambda = 0.001$, 
parameters $\beta1 = 0.9$, and $\beta2 = 0.999$, we set
$k=10$ for calculating $\theta_{th}$.

Experiments on both dataset used gradient descent optimisation for 75k 
iterations, with a learning rate of 0.01 for generating explanations in the 
the counterfactual explainer (module 3).

\section{Results and Discussion}

This section is organized into two parts, first evaluation of the anomaly 
detection and second the results from the counterfactual explanations. 

\subsection{Results from Anomaly detection}

Two AE models were trained, one for each dataset (SKAB and industrial). 
The anomaly detection threshold was calculated on the validation set, as 
described in Section~\ref{sec: methods}. The trained models were then 
evaluated on their respective test sets. The performance of the anomaly 
detector is summarized in Table~\ref{tab:anomaly_detection}.

The SKAB dataset results show satisfactory performance with F1-score and 
Recall around 0.7, along with a False Positive Rate (FPR) of 0.24. 
The industrial dataset exhibits exceptional performance, achieving 
F1-score and Recall close to 0.9, with a perfect zero FPR. Anomaly 
detection confusion matrix for both datasets can be found in 
Appendix A1.

\begin{table}[h]
    \begin{center}  
		
    \caption{Evaluating anomaly detection models on SKAB and industrial dataset.}
    \label{tab:anomaly_detection}
    \scalebox{1}{ 
    \begin{tabular}{c||c|c|c}
        \textbf{Dataset} & \textbf{F1-score} & \textbf{Recall} & \textbf{FPR}\\
        \hline
         \hline

        \textbf{SKAB} &    0.68 &	0.72 &	0.24 \\
        
        \textbf{Industrial data} &   0.94 &	0.88 &	0 \\
        
    \end{tabular}
    }
    \end{center}

\end{table}

\subsection{Results from counterfactual Explanation}

To demonstrate the effectiveness of our method in explaining time-series anomalies, 
we compare it with two other approaches:

\begin{itemize}
    \item \textbf{Reconstruction}: This method directly uses the AE reconstruction 
    as the explanation for an anomaly. This is based on the assumption that the
    reconstructions are projected onto the normal space, hence, a plausible 
    counterfactual explanation.

    \item \textbf{Counterfactual Explainer} (Without Feature Selection): This 
    approach utilizes a counterfactual explainer (module 3) to generate 
    explanations directly for all features, similar to gradient-based 
    counterfactual explanations with  $\lambda=1$  in equation~(\ref{eq: couter_cost}). 
    This essentially explains every feature without any selection.

    \item \textbf{Our Proposed Approach} (With Feature Selection): This combines a 
    feature selector (module 2) and a counterfactual explainer (module 3). The feature 
    selector identifies the most relevant features, and the counterfactual explainer 
    then focuses its explanation on these selected features only, with $\lambda=0$  
    in equation~(\ref{eq: couter_cost}).
\end{itemize}

We evaluate the explanations generated by these three approaches using three metrics: 
validity, sparsity, and distance. These metrics are explained in detail in 
section~\ref{sec: explainability eval}. The results of this comparison are presented 
in Table~\ref{tab: explanations}.


\begin{table}[h]
    \begin{center} 
    \caption{Compilation of evaluation measures from SKAB and industrial 
    dataset. The arrow direction indicates if higher or lower values that 
    makes the approach better. }
    \label{tab: explanations}
    \scalebox{0.80}{ 
    \begin{tabular}{c|c|c|c|c}
        \textbf{Dataset} & \textbf{Method} & \textbf{Validity} $\uparrow$ & 
        \textbf{Sparsity} $\downarrow$ & \textbf{distance} $\downarrow$\\
        \hline
         \hline
        

        SKAB & Reconstruction & 1.0 &	1.0 &	0.246 \\
        SKAB & Counterfactual & 0.72 &	1.0 &	0.214 \\
        SKAB & \textbf{Ours} & 0.67 &	0.16 &	0.150 \\

        \hline
        \hline

        Industrial data & Reconstruction & 1.0 &	1.0 &	0.140\\
        Industrial data & Counterfactual  & 0.93 &	0.99 &	0.200 \\
        Industrial data & \textbf{Ours} & 0.99 &	0.17 &	0.156 \\

    \end{tabular}
    }

    \end{center}
\end{table}

Table~\ref{tab: explanations} shows that our approach has reasonably good
validity and distance values compared to the other two simpler methods, but with
a much better sparsity values than the other methods. Note that the
reconstruction method will always have the highest possible validity value 
due to its nature that the reconstructions are in the same
manifold as training data.
So this method scores best in this validity measure on both datasets. The counterfactual
explainer (without feature selection) has higher validity measure than our
proposed method on the SKAB data. The counterfactual explainer (without feature
selection) has an advantage of being able to vary all features to provide
explanations. This does not necessary mean that the explanation will be more
meaningful as by adjusting all features simultaneously the information 
(the reasons) about the raised anomaly gets diluted. Additionally, altering all signals by
the counterfactual explainer (without feature selection) results in a the
sparsity scores much worse than our method (with feature selection). Scores form
our approach are consistently good in all three measures. To look further into
the meaningfulness of the given explanations we illustrate some scenarios in
section~\ref{sec: plot_explain }.

\begin{figure}[h]
    \centering
    \includegraphics[width=0.5\textwidth]{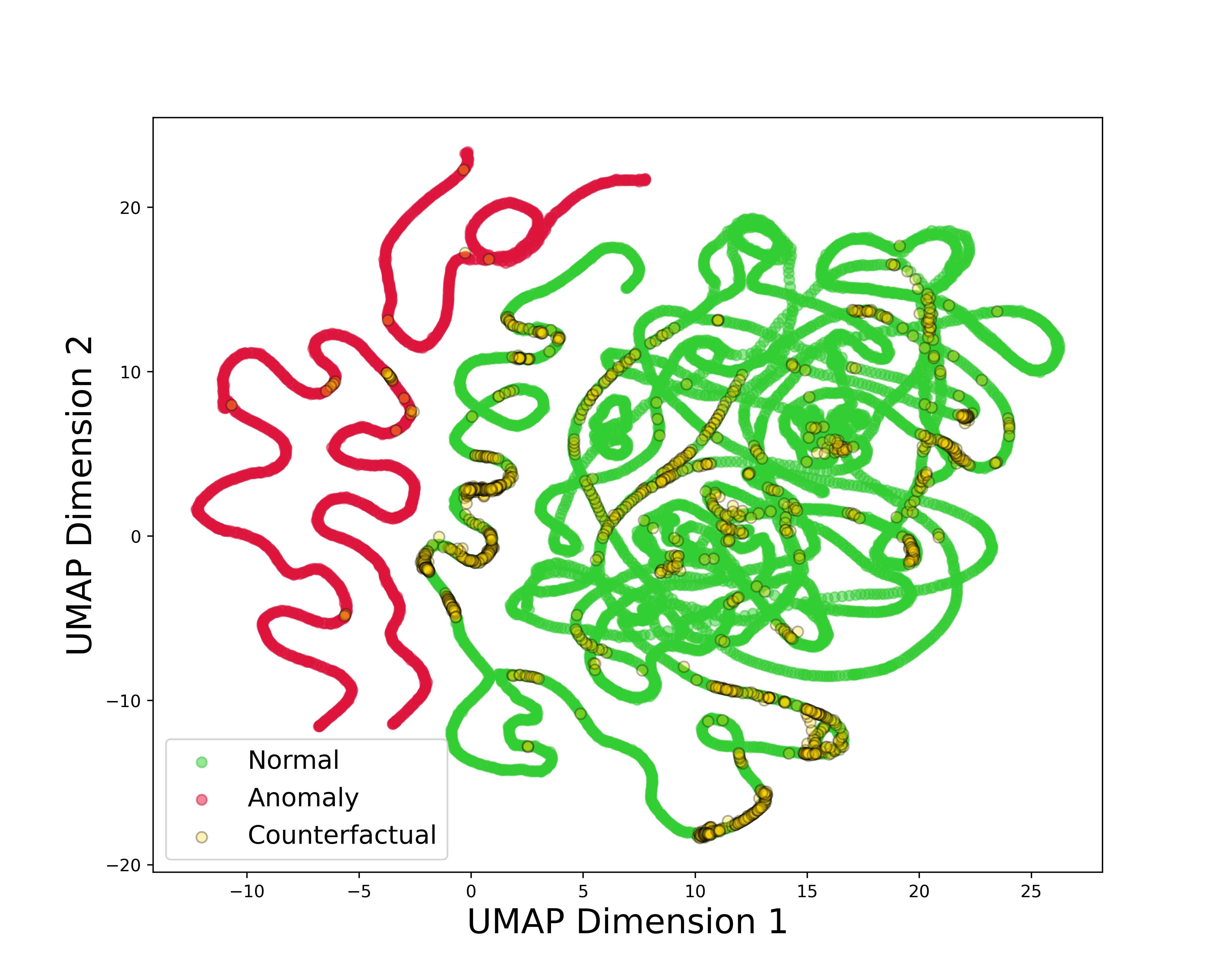}
    \caption{Industrial data: UMAP embedding learnt on no-fault and anomalous data 
    from the test set. Later the generated counterfactual is projected into the 
    same embedding.}
    \label{fig: UMAP}
\end{figure}

We leverage UMAP embedding (a dimension reduction technique) to achieve two
objectives: visualize the relationship between the generated counterfactuals and
the test data, and evaluate the \textit{validity} of the explanations independent of the
model used for counterfactual generation. In Figure~\ref{fig: UMAP} we visualize
the UMAP embedding trained on the test-set data from the industrial dataset.
Green points represent the non-faulty data (based on ground truth), red points
represent the anomalies (based on ground truth), and yellow points represent the
projected counterfactuals (generated form our approach). As evident from the
Figure~\ref{fig: UMAP} the majority of counterfactuals projected on top of the
green normal data embeddings, indicating that they represent valid non-faulty
behaviors. Only a few, 12 out of 1350 explanation are non-valid (which is
reflected in the \textit{validity} measure). These non-valid samples are projected onto
the same space as the red faulty data embedding. The lack of valid explanations can
be due to parameter selection, optimisation budgets and quality of the feature
selection.  The validity in confusion-matrix form for the SKAB test data is given in 
Table~\ref{tab: skab_valid_confusion} in Appendix~A2, the validity confusion-matrix 
form for real-world industrial test data is given in 
Table~\ref{tab: real_world_valid_confusion} in Appendix~A2.

\subsubsection{Plots showing insights on the explanations}
\label{sec: plot_explain }

\begin{figure}[t]
    \centering
    \includegraphics[width=0.47\textwidth]
    {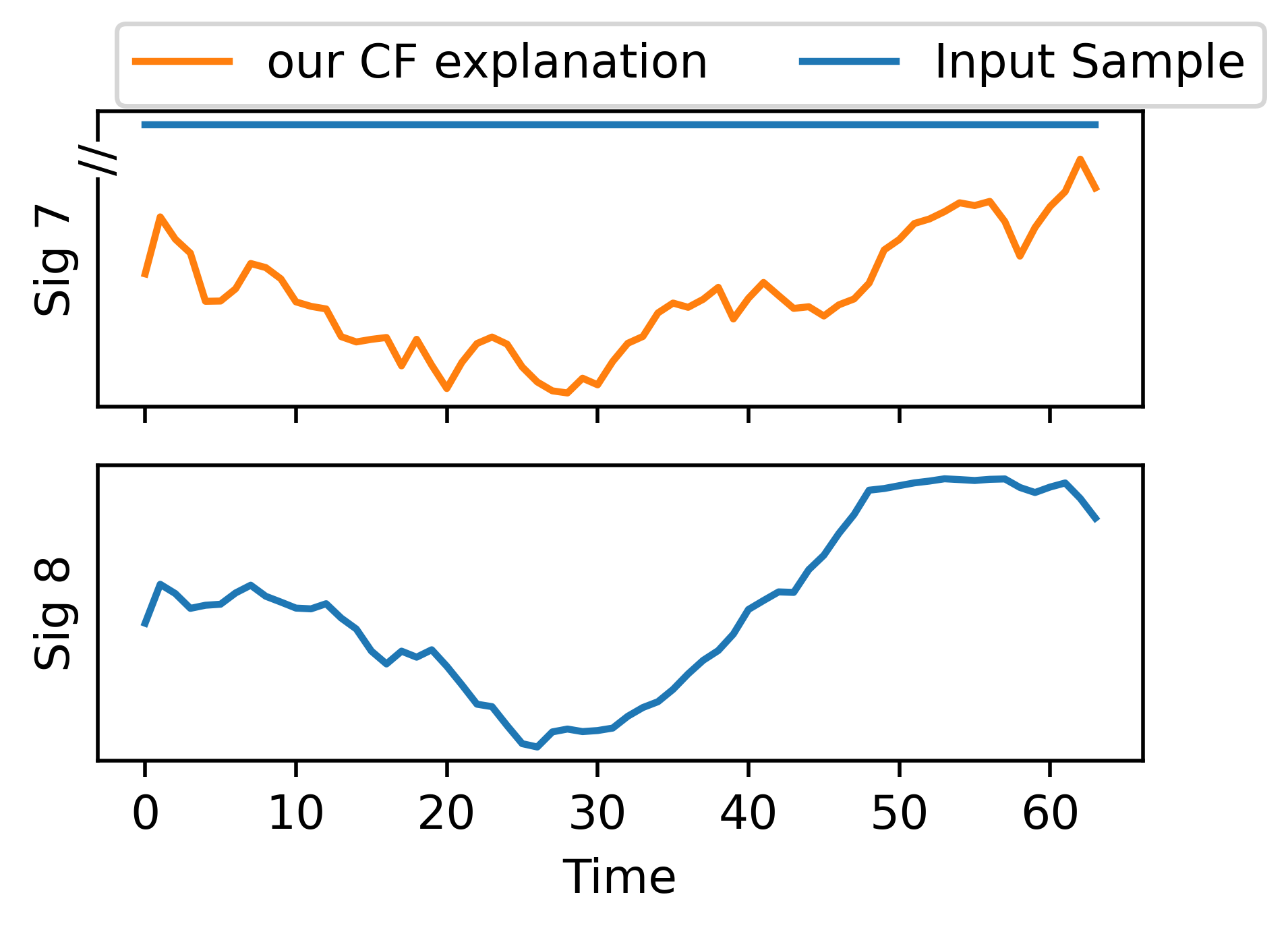}
    \caption{Plot of counterfactual explanation generated by our approach for 
    industrial dataset. This plotted sample was of correlation loss anomaly. 
    Signal 7 and signal 8 in blue show the input and signal 7 in orange shows 
    the explanation.}
    \label{fig: explanation_plot}
\end{figure}

In this section, we show two different explanation scenarios, one from the
industrial and the other from the SKAB dataset. Scenario 1 is from the
industrial dataset and is illustrated in the Figure~\ref{fig: explanation_plot}.
The time-window plotted in Figure~\ref{fig: explanation_plot} was classified as
anomalous and signal 7 was selected as high impact feature. In Figure~\ref{fig:
explanation_plot} we show the input signal 7 and signal 8 in blue and the
counterfactual explanation in orange (see Figure~\ref{fig: methods_comparision}
in the Appendix A3 for comparison with reconstruction and counterfactual signals). 
The root cause of this anomaly is a loss of correlation in signal 7. In normal (no-fault)
data signal 7 and signal 8 are correlated with a median correlation coefficient
of 0.99 and our explanation restored the correlation between the signals on the
anomalous data (of this type) to a median correlation coefficient of 0.93.

\begin{figure}[ht]
    \centering
    \includegraphics[width=0.47\textwidth]{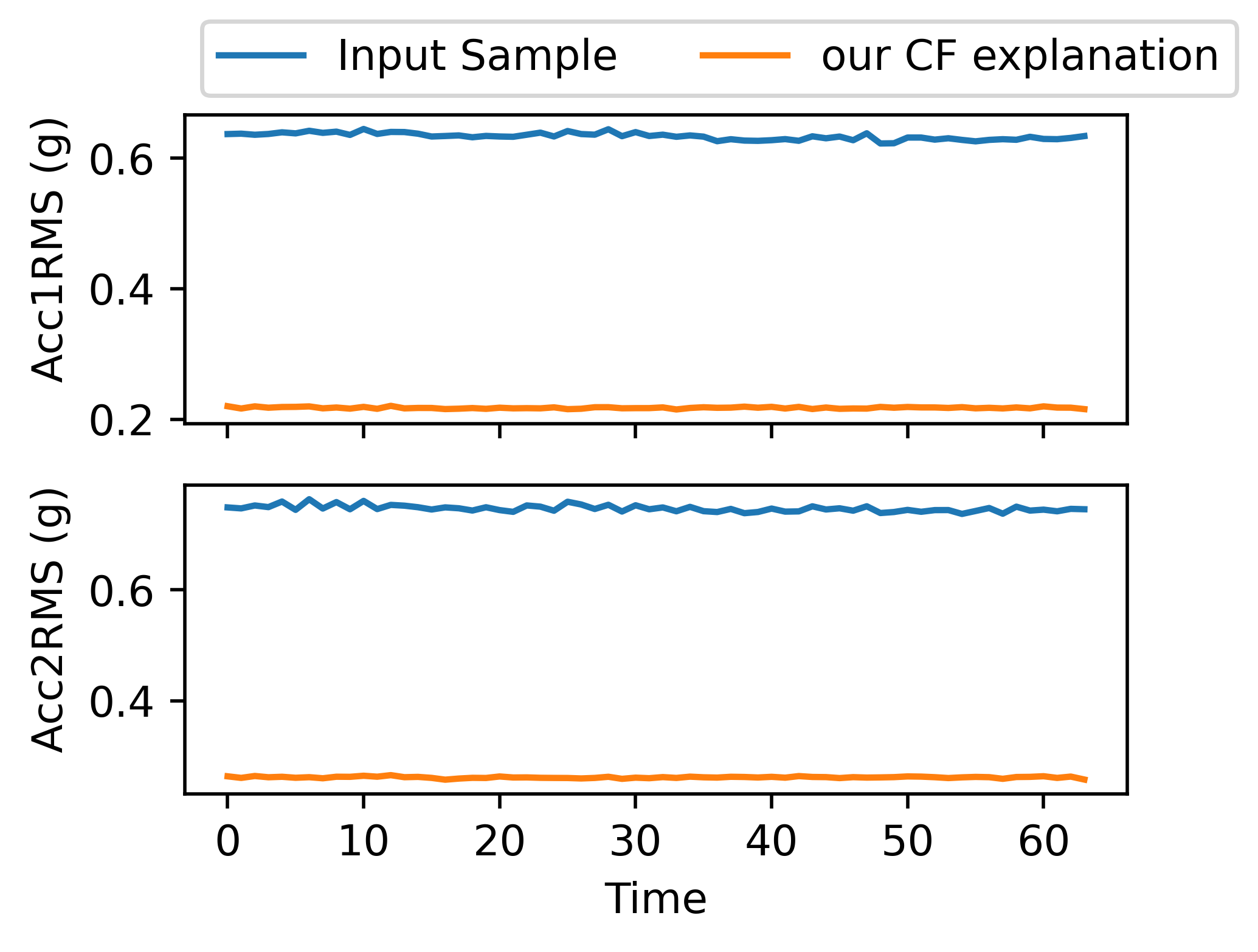}
    \caption{Plot showing the counterfactual explanations provided by our 
    approach and the anomalous samples. Only the high impact features that 
    were explained are plotted.}
    \label{fig: skab_explanation_plot}
\end{figure}

Figure~\ref{fig: skab_explanation_plot} shows the second scenario from SKAB
data. Here the selected anomalous window belongs to the rotor imbalance anomaly.
This window was classified as anomaly and our feature selector selected Acc1RMS
and Acc2RMS signals which belong to the accelerometer sensors as high impact
features. The explanation from our approach indicates that the vibrations
observed by the accelerometer should be lower to be classified as normal (see
the Figure~\ref{fig: scania_explains} in Appendix A3 to see the comparison with
CF and reconstruction signals). 

In Scenario 1, the explanation hints that the the correlation between signal 7
and signal 8 is broken by the flat line and is confirmed by the correlation
analysis. Combining this explanation with the domain expertise, it is easy to
conclude that the sensor for signal 7 is broken. In Scenario 2 from the
explanation we know that we have too high vibrations that often are originated
by rotor imbalance. The explanations provided by our approach are
meaningful in the context of system functionality and provides insights about
the nature of the anomaly when compared to other approaches. This is due to it's 
capacity to select features for explanation. The comparison between different 
approaches can be seen in the detail in the Figure~\ref{fig: scania_explains} 
and Figure~\ref{fig: methods_comparision} provided in Appendix A3.

\section{Conclusion}
In summary, our work proposes a method for explaining AE-based anomaly detection
for time-series data, based on relevant feature selection and counterfactual 
explanations. This approach can answer on which features the anomaly is located
together with why the sample was classified as an ano-maly. We find that these
explanations have consistently good scores in all three measures,
\textit{validity}, \textit{sparsity} and \textit{distance}, which translates
into useful and actionable insights from a diagnostic perspective. We give two
examples, one from a benchmark dataset and one from an industrial dataset, on
how the proposed method can help to diagnose the classified anomalies from the
AE anomaly detection model. This contribution serves as a diagnostic tool,
enhancing our understanding and analysis of anomalous events. Note that the 
quality of explanation depends on the performance of the selected anomaly 
detection model, parameter selection and the quality of feature selection.


Future work can focus on different optimisations for the explanation, improve 
the quality of the feature selector and understand the model relation with the 
explainer.

\section*{Acknowledgment}
The project was funded FFI Vinnova under project number 2020-05138. We thank Deepthy and Swathy for good discussions.

\bibliographystyle{IEEEtran}
\bibliography{IEEEabrv,references}

\section*{Appendix}

\subsection*{A1. Confusion Matrix for the Anomaly detector}

In this section, the confusion matrices for the anomaly detector on SKAB 
and real-world industrial  dataset are presented in Table~\ref{tab: SKAB_confusion} 
and Table~\ref{tab: real_world_confusion}, respectively.


\begin{table}[h]
    \centering

    \caption{Confusion Matrix for SKAB test data. }
    \label{tab: SKAB_confusion}

\begin{tabular}{c > {\bfseries}r @{\hspace{0.7em}}c @{\hspace{0.4em}}c @{\hspace{0.7em}}l}
  \multirow{10}{*}{\rotatebox{90}{\parbox{1.1cm}{\bfseries\centering Actual\\ value}}} & 
    & \multicolumn{2}{c}{\bfseries Prediction outcome} & \\
  & & \bfseries P & \bfseries N & \bfseries Total \\[0.5em]
  & \bfseries P$'$ & 2788 & 1088 & 3876 \\[0.5em]
  & \bfseries N$'$ & 1573 & 4977 & 6550 \\[0.5em]
  & \bfseries Total & 4361 & 6065 & 10426
\end{tabular}
\end{table}

\begin{table}[h]
    \centering

    \caption{Confusion Matrix for real-world industrial test data. }
    \label{tab: real_world_confusion}

\begin{tabular}{c > {\bfseries}r @{\hspace{0.7em}}c @{\hspace{0.4em}}c @{\hspace{0.7em}}l}
  \multirow{10}{*}{\rotatebox{90}{\parbox{1.1cm}{\bfseries\centering Actual\\ value}}} & 
    & \multicolumn{2}{c}{\bfseries Prediction outcome} & \\
  & & \bfseries P & \bfseries N & \bfseries Total \\[0.5em]
  & \bfseries P$'$ & 1350 & 171  & 1521  \\[0.5em]
  & \bfseries N$'$ & {0} & 2834 & 2834\\[0.5em]
  & \bfseries Total & 1350  & 3005 & 4355
\end{tabular}

\end{table}

\subsection*{A2. Confusion Matrix like expression for validity using our approach}

In this section, we show valid samples in a confusion-matrix like setting for 
SKAB and real-world industrial  dataset are presented in Table~\ref{tab: skab_valid_confusion}  
and Table~\ref{tab: real_world_valid_confusion}  respectively.

\begin{table}[h]
    \centering

    \caption{Validity confusion Matrix for SKAB test data. }
    \label{tab: skab_valid_confusion}

\begin{tabular}{c > {\bfseries}r @{\hspace{0.7em}}c @{\hspace{0.4em}}c @{\hspace{0.7em}}l}
  \multirow{10}{*}{\rotatebox{90}{\parbox{1.1cm}{\bfseries\centering Model\\ Prediction}}} & 
    & \multicolumn{2}{c}{\bfseries Prediction outcome} & \\
  & & \bfseries Valid & \bfseries Not Valid & \bfseries Total \\[0.5em]
  & \bfseries True Positives & 1885 & 903  & 2788  \\[0.5em]
  & \bfseries False Positives & 1068 & 505 & 1573\\[0.5em]
  & \bfseries Total & 2953  & 1048 & 4361
\end{tabular}

\end{table}

\begin{table}[h!]
    \centering

    \caption{Validity confusion Matrix for real-world industrial test data. }
    \label{tab: real_world_valid_confusion}

\begin{tabular}{c > {\bfseries}r @{\hspace{0.7em}}c @{\hspace{0.4em}}c @{\hspace{0.7em}}l}
  \multirow{10}{*}{\rotatebox{90}{\parbox{1.1cm}{\bfseries\centering Model\\ Prediction}}} & 
    & \multicolumn{2}{c}{\bfseries Prediction outcome} & \\
  & & \bfseries Valid & \bfseries Not Valid & \bfseries Total \\[0.5em]
  & \bfseries True Positives & 1338 & 12  & 1350  \\[0.5em]
  & \bfseries False Positives & {0} & {0} & {0}\\[0.5em]
  & \bfseries Total & 1338  & 12 & 1350
\end{tabular}

\end{table}

\subsection*{A3. Plot comparing different approaches}

\begin{figure}[h]
    \centering
    \includegraphics[width=0.4\textwidth]{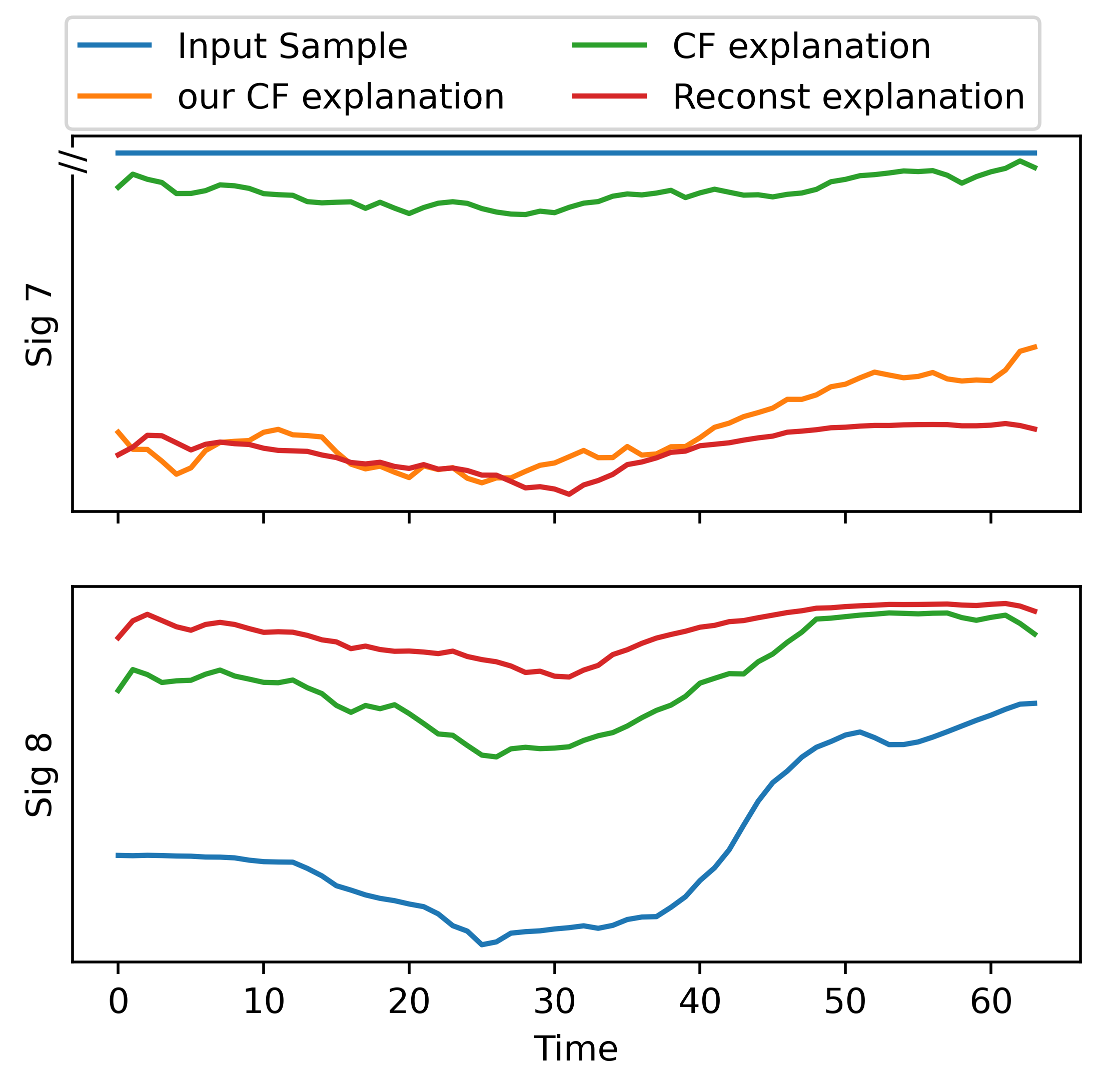}
    \caption{Plot showing the explanations provided by reconstruction, 
    counterfactual(CF) based (i.e., without feature selector) and our 
    approach (i.e., with feature selector). Additionally the input sample 
    is plotted.}
    \label{fig: scania_explains}
\end{figure}

\begin{figure}[h]
    \centering
    \includegraphics[width=0.45\textwidth]{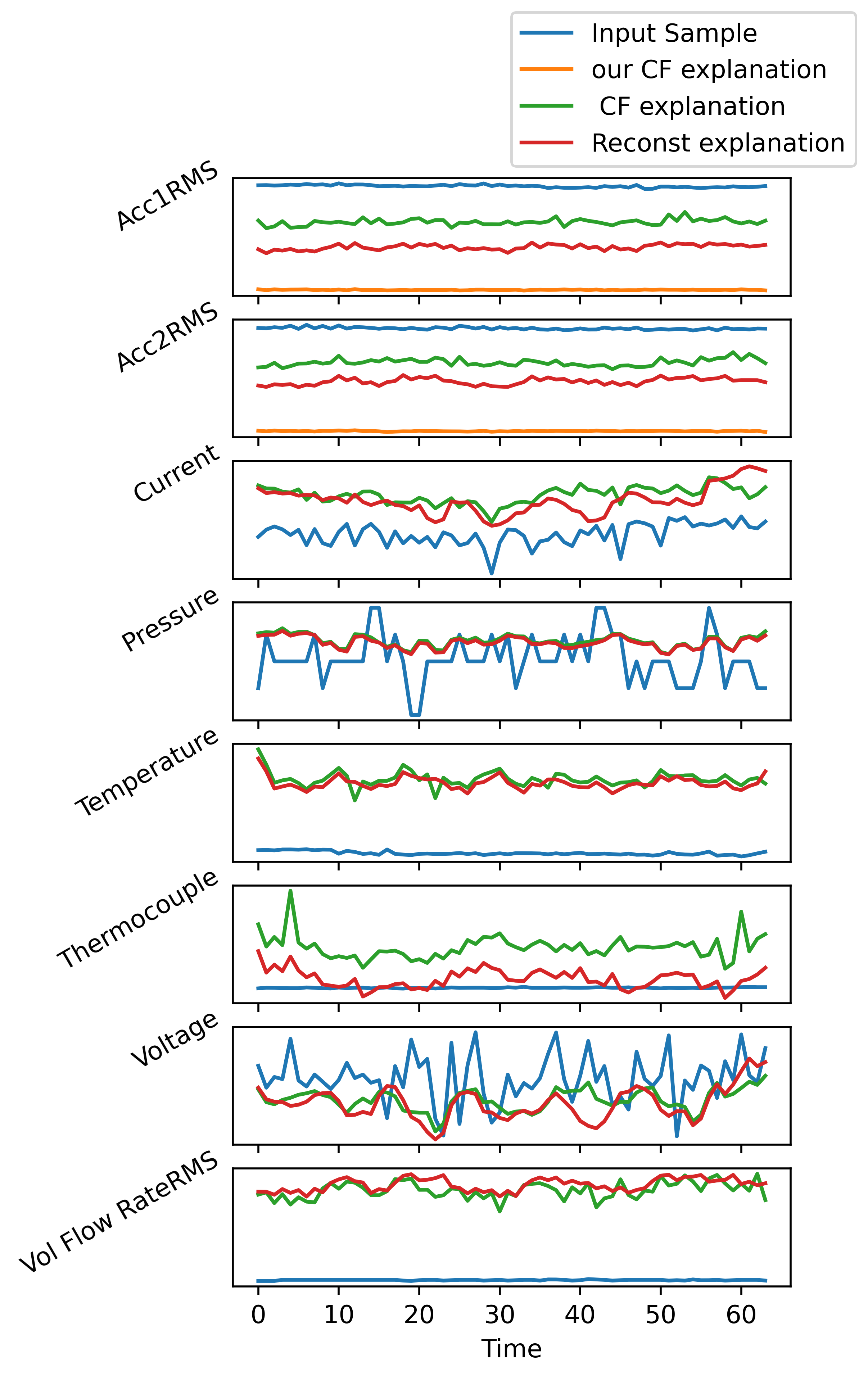}
    \caption{Plot showing the explanations provided by reconstruction, 
    counterfactual(CF) based (i.e., without feature selector) and our approach 
    (i.e., with feature selector). Additionally the input sample is plotted. }
    \label{fig: methods_comparision}
\end{figure}

A sample from rotor-imbalance anomaly is plotted along with different explanations 
in the  Figure~\ref{fig: methods_comparision}. The plotted sample  is the same as 
in the Figure~\ref{fig: skab_explanation_plot}. In figure \ref{fig: methods_comparision}, 
explanations from different methods are compared. It can be seen that other approaches 
explains by changing all the features where as the explanation from our approach 
changes only \textit{ACC1RMS} and \textit{ACC2RMS} signals. In similar way , for 
the sample plotted in the Figure~\ref{fig: explanation_plot}, in 
Figure~\ref{fig: scania_explains}, we compare our approach with other type of explanations.

\clearpage

\end{document}